\documentclass[sigconf]{acmart}

\AtBeginDocument{%
  }

\copyrightyear{2024}
\acmYear{2024}
\setcopyright{rightsretained}
\acmConference[DocEng '24]{ACM Symposium on Document Engineering 2024}{August 20--23, 2024}{San Jose, CA, USA}
\acmBooktitle{ACM Symposium on Document Engineering 2024 (DocEng '24), August 20--23, 2024, San Jose, CA, USA}
\acmDOI{10.1145/3685650.3685673}
\acmISBN{979-8-4007-1169-5/24/08}

\begin{document}

\title{Detecting AI-Generated Texts in Cross-Domains \\
}

\author{You Zhou}
\affiliation{%
  \institution{Miner School of Computer \& Information Sciences \\University of Massachusetts, Lowell, MA, USA}
  \city{}
  \state{}
  \country{}
}
\email{you\_zhou1@student.uml.edu}

\author{Jie Wang}
\affiliation{%
  \institution{Miner School of Computer \& Information Sciences \\University of Massachusetts, Lowell, MA, USA}
  \city{}
  \state{}
  \country{}
}
\email{jie\_wang@uml.edu}


\begin{abstract}
Existing tools to detect text generated by a large language model (LLM) have met with certain success, but their performance can drop when dealing with texts in new domains. To tackle this issue, we train a ranking classifier called RoBERTa-Ranker, a modified version of RoBERTa, as a baseline model using a dataset we constructed that includes a wider variety of texts written by humans and generated by various LLMs. We then present a method to fine-tune RoBERTa-Ranker that requires only a small amount of labeled data in a new domain. Experiments show that this fine-tuned domain-aware model outperforms the popular DetectGPT and GPTZero on both in-domain and cross-domain texts, where AI-generated texts may either be in a different domain or generated by a different LLM not used to generate the training datasets. This approach makes it feasible and economical to build a single system to detect AI-generated texts across various domains.
\end{abstract}




\maketitle

\section{Introduction}
AI-generated texts could be misused for foul purposes, such as cheating on assignments or spreading misinformation from LLM hallucinations \cite{huang2023survey}. 
Detection tools have been constructed, such as
DetectGPT \cite{mitchell2023detectgpt} and GPTZero \cite{gptzero}. 
At the time of writing this paper, there is no single model that works well in detecting AI-generated texts across domains. To add to the difficulty of this task, we note that AI-generated texts may originate from different LLMs and prompts, with various generation configurations and techniques, possibly even involving post-generation human modifications.

To address this issue, we present a new approach by first training a detection model as a baseline with a large labeled dataset consisting of texts in a wide variety of domains, written by human writers and generated by various LLMs.
The baseline model we choose is a modified version of RoBERTa \cite{liu2019roberta}, called RoBERTa-Ranker, which is
a ranking classifier. The modification is twofold: (1) Replace the default loss function with the margin ranking loss function, which enhances its robustness for imbalanced data. (2) Add a mean-pooling layer to the encoding layer to extract features from the input text. 
RoBERTa-Ranker on an input article outputs a label with a confidence level for the purpose of ranking, where the label represents either human or AI.

RoBERTa-Ranker achieves high F1 scores on texts in the same domain as the training dataset. 
While retraining RoBERTa-Ranker on a large, domain-specific dataset can improve cross-domain accuracy, this approach is expensive and may not always be feasible, as such datasets are hard to come by and training a new model takes excessive time. To overcome this obstacle, we devise a fine-tuning method that requires only a small labeled dataset in a new domain.
Such a dataset can be constructed by selecting a small number of human-written articles in the new domain from web publications and then adding AI-generated texts via rewriting or generating articles on the same topics.
Denote by FT-RoBERTa-Ranker a fine-tuned RoBERTa-Ranker on such a dataset in the new domain.
%
Experiments show that FT-RoBERTa-Ranker is time-efficient to fine-tune and performs well across various domains, outperforming both DetectGPT and GPTZero.

We can therefore combine the fine-tuned models for cross domains to form a single multi-modal system to detect LLM-generated texts of various kinds. When a new domain of texts is encountered, we can simply add a fine-tuned  model for this domain to the system.

\section{Previous approaches}
\label{sec:related work}

Existing methods for detecting AI-generated texts are either black-box detection or white-box detection\cite{tang2023science}. The former treat an LLM as a black box with only API-level access to LLMs, relying on collecting samples from both human-written and AI-generated texts to train a classifier\cite{quidwai2023black}. These methods perform well when AI-generated texts exhibit certain linguistic or statistical patterns. As LLMs evolve and improve, however, traditional binary classification methods have become less effective. White-box detection\cite{sadasivan2024aigenerated} offers an alternative but requires full access to LLMs, which only works for open-source LLMs. Various contextual properties have been used to analyze linguistic patterns in human-written and AI-generated texts, such as vocabulary, part-of-speech, dependency parsing, sentiment analysis, and stylistic features. There are also statistical methods based on linguistic feature statistics \cite{gallé2021unsupervised}. Linguistic features such as branch properties observed in text syntax analysis, function word density, and constituent length have also been used. DetectGPT and GPTZero are popular detection tools among the existing systems. DetectGPT is an unsupervised method that detects AI-generated text by generating random perturbations from an LLM and then obtaining the probabilities of the original and perturbed texts from the LLM likely to have generated the text. GPTZero takes a different approach, combining various techniques such as training on labeled data and analyzing the text itself to make a judgment.

\section{Our approach}
\label{sec:model}

Recent studies have confirmed the outstanding performance of fine-tuned models in the BERT family on detecting AI-generated texts \cite{Gambini2022OnPD}, achieving an average accuracy of 95\% for texts in the test set following the standard training-test split, outperform zero-shot and watermarking methods, and exhibit some resilience to various attack techniques. 
%
%
This motivates us to modify a model in the BERT family as a baseline model.

\subsection{RoBERTa-Ranker, the baseline}

RoBERTa-Ranker, modifying RoBERTa, uses the margin ranking loss function to enhance robustness on imbalanced data, rather than 
a conventional loss function such as Binary Cross-Entropy Loss typically used in
a binary classifier.
Margin Ranking Loss is used 
to compare the score differences between two samples, 
ensuring that the distance between samples of the same class is greater than the margin, while the distance between samples of different classes is smaller than the margin. 
%
We also add a mean pooling layer on top of RoBERTa's encoding layer to extract features from the input text, converting variable-length text sequences into fixed-length vector representations. This vector captures general information within the input sequence, which is then passed as input to subsequent neural network layers for classification.


Training is carried out over the dataset constructed in Section \ref{sec:dataset} with the
standard 80-20 split of training-test data.
Experiments show that the baseline RoBERTa-Ranker performs well on texts in the test set. 

\subsection{FT-RoBERTa-Ranker for cross domains}
\label{sec:ft}

To achieve better performance on cross-domain texts and avoid retraining the model on a large labeled dataset for the new domain, we devise a method that only needs 
a very small labeled dataset to fine-tune RoBERTa-Ranker to produce FT-RoBERTa-Ranker

Let $n > 0$ be a small integer (for example, $n = 1,000$). We first select $n$ human-written articles in the new domain from arXiv, PUBMED, or other public sources. We then use LLMs to 
rewrite a significant portion of an article or the entire article to generate $n$ articles. We now have a labeled dataset $S$ for the new domain, with $n$ articles labeled with Human and $n$ articles with LLM.
We use a 50-50 split to select $n$ articles for fine-tuning and the rest for testing.
For each article in $S$, the dataset for fine-tuning, we first use the baseline RoBERTa-Ranker to predict a label and a confidence level. Note that a predicted label may or may not be the same as the true label.
The objective of fine-tuning is to achieve the highest accuracy possible.

To do so, for each article $A \in S$, we calculate its similarity with the rest of the articles in $S$ using embedding similarity and linguistics features statistics, 
including lexical features, readability score, diversity and richness of vocabulary. We normalize the feature scores and sum them up as the final similarity score. We then select $k$ articles with a predicted label of Human with the highest confidence levels, denoted by $S_H$. Likewise, we select $k$ articles with a predicted label of LLM with the highest confidence levels, denoted by $S_M$. Note that the accuracy of the baseline RoBERTa-Ranker on a new domain can at least reach a certain positive percentage (e.g., 70\%), ensuring a sufficiently large number $K$ of articles with either predicted label. We then choose a smaller value for $k$ with $k < K$. For example, choosing $k = 30$ would be reasonable. Thus, training a solid baseline model is crucial in this approach.

Next, we compute the average similarity of $A$ to $S_H$ and $S_M$, respectively, denoted by $avgH_A$ and $avgM_A$. Let $r_A = avgH_A/avgM_A$, called the human-LLM similarity ratio.
Suppose the predicted label of $A$ is Human, then $r_A$ is ``true positive" if the true label of $A$ is Human, and ``false positive" otherwise.
We use the ROC curve analysis to identify the point on the ROC curve,
denoted by $\delta$, such that the F1 score is maximized. We use $\delta$ as the threshold for this domain, which 
%
represents the best trade-off between precision and recall for the classification task.




After fine-tuning, for each article $A'$ in the same domain but not in the training set $S$,
compute $r_{A'}$ and label $A'$ as human-written if $r_{A'} \geq \delta$ and AI-generated otherwise.


\section{Evaluation}
\label{sec:experiments}

Experiments were carried out on a GPU server with a 3.60 GHz 8-core Intel Core i7-9700K CPU and an NVIDIA RTX A6000 GPU with 64 GB of RAM. 

\subsection{Dataset}
\label{sec:dataset}

The dataset we constructed is called LLMCheck. It first combines existing datasets to form a base dataset, and
then expands it using various LLMs to generate new texts.
The existing datasets are TuringBench, TweekFake, and PERSUADE 2.0 (PERSUADE for short).

TuringBench \cite{uchendu2021turingbench} is a collection of 10K human-authored news articles, primarily from authoritative sources such as CNN, with each article ranging from 200 to 400 words. AI-generated texts in this dataset were produced by 19 different text generation models popular at the time, 
aiming at exploring the "Turing Test" challenge. 

TweepFake \cite{Fagni_2021} is a collection of tweets from both genuine and fake accounts, comprising a total of 25,836 tweets by 23 bots and 17 human accounts, with equal numbers of samples written by humans and generated by machines. The machine-generated tweets were produced using various techniques, including GPT-2, RNN, Markov, LSTM, and CharRNN.
To strengthen the dataset of short messages like tweets, we also incorporate question-answering pairs (QAPs) from
%
HC3 \cite{guo2023close}, which consists of over 37K QAPs covering various domains, including computer science, finance, medicine, law, and psychology, among others. It contains answers from both humans and ChatGPT 3.5 for the same questions. 
But the ChatGPT-generated answers were produced via homogeneous prompts that lack the desired diversity.

PERSUADE \cite{persaude_corpus} is a collection of over 25K persuasive essays written by US students in grades 6 to 12, featuring independent writing and resource-based writing on 15 topics. Each essay has an overall rating and a proficiency score for each argument and discourse element.

LLMCheck was constructed using diverse configurations and prompting strategies to promote diversity and complexity in the generated texts. Texts are generated using a combination of  temperatures and top-k values; prompts with randomly masked words in text; reconstruction of the original text; prompts without using the source text; prompts using the source text; and prompts to completely rewrite text. As a result, LLMCheck contains the following kinds of texts:
Articles suitable for readers with different educational backgrounds in terms of age groups or grade levels;
articles generated using different prompting methods;
articles with the first half being human-written or generated by an LLM and the second half being human-written or generated by a different LLM; and
articles generated by various LLMs, including GPT-3.5, GPT-4, and LLaMA-7B.
LLMCheck is available at https://github.com/zyloveslego/LLMCheck.

\subsection{Results and analysis}

We train RoBERTa-Ranker on LLMCheck with the standard 80-20 training-test split. Its performance is evaluated on the test set. Note that HC3 consists of short texts similar to TweepFake, so evaluation on TweepFake is deemed sufficient. We then assess the fine-tuned RoBERTa-Ranker on texts from cross domains by constructing small datasets in three new domains. 

%


\subsubsection{In-domain}
\label{sec:in-domain}

Table \ref{table:1} shows the F1 scores of RoBERTa-Ranker
on the in-domain test dataset, as well as the F1 scores of DetectGPT and GPTZero
on the same test dataset. RoBERTa-Ranker outperforms both DetectGPT and GPTZero on the in-domain test dataset, with satisfactory F1 scores ranging from 92.1\% to 97.3\%. Note that the low F1 scores for DetectGPT on all test datasets are likely attributed to the LLM models used to generate the text, which do not provide token probabilities needed by DetectGPT.

\begin{table}[h]
\caption{F1 scores on in-domain test datasets (\%) }
\begin{tabular}{l|c|c|c}
\hline
          & TuringBench & TweepFake & PERSUADE \\ \hline                                                                            
DetectGPT & 68.5        & 60.7      & 72.4      \\ \hline
GPTZero   & 81.2       & 89.4      & 94.6      \\ \hline
RoBERTa-Ranker          & \textbf{88.1} & \textbf{92.1}      & \textbf{97.3}  \\ \hline
\end{tabular}
\label{table:1}
\end{table}

%


\subsubsection{Out of domain}

To evaluate the performance of the fine-tuned RoBERTa-Ranker on out-of-domain texts, we constructed three small datasets in different domains, each consisting of 1,000 articles. 
They are (1) a set of the abstracts of research papers selected at random from arXiv.org, published before 2022 to ensure that
they do not include texts generated by GPT-3 or above;
(2) a set of stories selected at random from the Story Gene \cite{fan2018hierarchical}, a collection of 300K human-written stories paired with writing prompts;
(3) a set of essays selected at random from the human-written essays in PERSUADE 2.0.

We then used GPT-4 with a diverse set of prompts to generate 1,000 abstracts from each of the selected human-written abstracts and
1,000 stories from each of the selected human-written stories.
Since GPT-4 was used to generate LLMCheck, we use three different LLMs to generate
1,000 essays from each of the selected human-written essays. These LLMs are Claude 3 Sonnet,
Moe 8x7B, and Gemma-7B, which, generate, respectively, 333, 333, 334 essays.
%
We also used prompts  
in the WritingPrompts dataset to direct GPT-4 to generate stories.

\begin{figure}[h]
    \centering
    \includegraphics[scale=0.4]{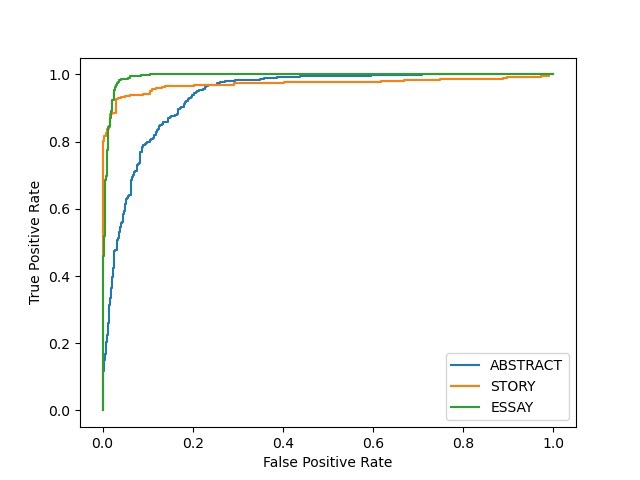}
    \caption{ROC curves for the H-LLM ratio}
    \label{fig:roc}
\end{figure}

We combined human-written and AI-generated texts to form three datasets of 2,000 texts each, called ABSTRACT, STORY, and ESSAY. We then fine-tuned RoBERTa-Ranker with a 50-50 split to obtain three threshold values for each dataset using the training dataset as discussed in Section \ref{sec:ft}, and used it to determine the label of an article in the test dataset for each domain. Figure \ref{fig:roc} depicts the ROC curve, which provides the threshold values of 0.457, 0.550, and 0.710 for ABSTRACT, STORY, and ESSAY, respectively.

While training the baseline RoBERTa-Ranker took more than 20 hours, fine-tuning an FT-RoBERTa-Ranker on a dataset in a cross domain of 1,000 articles only took several minutes to complete. 

Table \ref{table:2} shows the F1 scores of the fine-tuned RoBERTa-Ranker, denoted by FT-RoBERTa-Ranker,
as well as the F1 scores of DetectGPT and GPTzero.

\begin{table}[htbp]
\centering
\caption{F1 scores on out-of-domain test datasets (\%)}
\begin{tabular}{l|c|c|c}
\hline
           & ABSTRACT  & STORY  & ESSAY  \\ \hline
DetectGPT  & 65.9 & 71.0 & 72.3 \\ \hline
GPTZero    & 82.4 & 92.7 & 95   \\ \hline
RoBERTa-Ranker  & 78.3 & 93.2  & 96.9 \\\hline
FT-RoBERTa-Ranker & \textbf{85.8} & \textbf{94.0} & \textbf{97.1}  \\ \hline
\end{tabular}
\label{table:2}
\end{table}


FT-RoBERTa-Ranker improves the detection accuracy of the baseline model for each cross domain and outperforms both DetectGPT and GPTZero, with satisfactory F1 scores ranging from 85.8\% to 97.1\%. The reason for DetectGPT's low F1 scores is the same
as discussed in Section \ref{sec:in-domain}
Note that for both STORY and ESSAY, they are somewhat similar to the in-domain training datasets of TweepFake and PERSUADE. Stories are similar to tweets, and human-written essays are the same across domains. This explains why the F1 scores of FT-RoBERTa-Ranker for these domains are close to those of the baseline model.
%



\begin{figure}[h]
\centering
\includegraphics[width=\columnwidth]{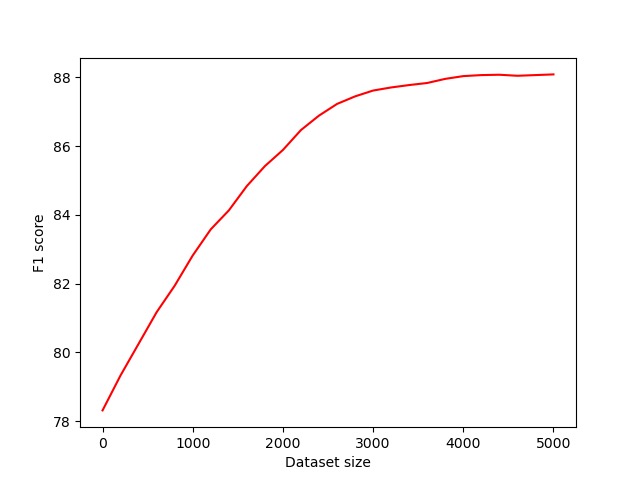}
\caption{F1 scores on datasets of different sizes}
\label{fig:2}
\end{figure}

For ABSTRACT, we note that abstracts of academic papers are substantially different from the training data, making it a truly new domain. In this case, FT-RoBERTa-Ranker for this new domain has shown substantial improvement. On the other hand, the F1 score for FT-RoBERTa-Ranker for ABSTRACT is still below 90\%, partly due to the baseline model not being trained with similar texts.
Increasing the size of the training datasets to fine-tune the baseline model is expected to help elevate accuracy. To verify this, we incrementally increased the size of the training dataset from 200 to 5,000 abstracts, constructed in the same way as described above, with an increment of 200 abstracts each time. Figure \ref{fig:2} shows the results, which confirm our expectation. Specifically, the F1 scores improve rapidly from about 78\% to about 88\% with datasets of sizes up to 3,000. Beyond that, the F1 scores continue to improve, but at a much slower pace.
The fine-tuning time is still in minutes. The longest is for the dataset of 5,000 abstracts, which is less than 20 minutes.

\subsection{Remark on domain determination}

To use FT-RoBERTa-Ranker for a specific domain, we need to know which domain an article belongs to so that an appropriate fine-tuned model can be chosen. This poses no problem for most applications. In the case where the underlying domain of an article is unknown, we propose the following method to determine its domain:
\begin{enumerate}
\item For each article in the small training set for each domain (assuming we have a pool of FT-RoBERTa-Rankers for various domains), compute a 10-dimensional vector of the content significance distribution of the first kind (CSD-1) for the article (see \cite{zhou2023content} for details), which has been shown to be a good representation of article types.

\item Compute the geometric center of these vectors to serve as the characteristic vector (CV) for the domain.

\item For the given article, compute its 10-dimensional vector of CSD-1. The domain with the nearest CV is the domain it belongs to.
\end{enumerate}

\section{Conclusion}
\label{sec:conclusion}
This paper proposes a systematic approach to constructing a single system to detect AI-generated texts cross domains: Train a solid baseline model on a large dataset of texts in a few domains, then fine-tune the baseline model for a new domain using a lightweight method and add it to the system. Experiments showed that this approach is efficient and improves detection accuracy.
%
%
%
%
We are interested in exploring the following topics for further investigations:
\begin{enumerate}
\item Conduct experiments and evaluations on more domains. 
\item Conduct comparisons with other exiting AI-generated text detection systems.
\item Develop a better baseline model and construct a more comprehensive dataset for training.  \item Develop a method to provide explanations for its classification decisions, allowing human users to evaluate the system's decisions.
\end{enumerate}

\bibliographystyle{ACM-Reference-Format}
\bibliography{mybibliography}

\end{document}